# Multi-Subspace Neural Network for Image Recognition

Chieh-Ning Fang, Chin-Teng Lin, *Fellow, IEEE*

*ABSTRACT—* In image classification task, feature extraction is always a big issue. Intra-class variability increases the difficulty in designing the extractors. Furthermore, hand-crafted feature extractor cannot simply adapt new situation. Recently, deep learning has drawn lots of attention on automatically learning features from data. In this study, we proposed multi-subspace neural network (MSNN) which integrates key components of the convolutional neural network (CNN), receptive field, with subspace concept. Associating subspace with the deep network is a novel designing, providing various viewpoints of data. Basis vectors, trained by adaptive subspace self-organization map (ASSOM) span the subspace, serve as a transfer function to access axial components and define the receptive field to extract basic patterns of data without distorting the topology in the visual task. Moreover, the multiple-subspace strategy is implemented as parallel blocks to adapt real-world data and contribute various interpretations of data hoping to be more robust dealing with intra-class variability issues. To this end, handwritten digit and object image datasets (i.e., MNIST and COIL-20) for classification are employed to validate the proposed MSNN architecture. Experimental results show MSNN is competitive to other state-of-the-art approaches.

# I.  INTRODUCTION

In image classification, the major challenge comes from intra-class variability [1], [2]. The intra-class variability issues arise from intensity of illumination, translation, scaling or rotation of photo. For traditional pattern recognition, feature extractor is designed to overcome intra-class variability and many experts in various domains are required to explore the suitable feature extractors. For example, scale invariant feature transform (SIFT) had been proven to effectively extract local image features for object recognition [3] and facial features drawn by kernel principal component analysis (Kernel PCA) produce good a performance for face recognition [4]. Despite some tasks applied with hand-crafted features gain promising achievements, hand-crafted feature extractors are usually constrained in specific applications.

Recently, deep learning draws lots of attention. Features could be automatically learned from data itself. The concept of deep learning is to generate different representations of data. By stacking similar structure, we expect to derive high-level and abstract meanings of data so the network should be more robust and invariant to deal with intra-class variability problem [5]. One of well-known deep learning algorithms is deep neural network (DNN). The performance of DNN was empirically worse than of shallow network with two or three hidden layers in 1992 [6]. The researchers inferred that initializing parameters near poor region makes DNN trained by gradient-based optimization algorithm get stuck in the bad classification performance [7]. Until 2006, Hinton et al had a purposed learning parameters to conquer the bad performance [8]. A greedy layer-wise unsupervised learning algorithm was introduced to deep belief network (DBN), restricted Boltzmann machine (RBM) is used for building blocks. Numerous studied about deep learning have been arose since then.

One of major improvement in image classification is made by biologically inspired deep architecture, convolutional neural network (CNN). Lecun presented CNN in 1995 [6] , and this algorithm has achieved remarkable records on various visual tasks such as handwritten digit recognition [6], large scale visual recognition [9], face recognition [10], etc. Two important factors in CNN are receptive field and shared weights. The receptive field was first presented in the experiment of the cat's visual system [11]. With local receptive fields, each single neuron is able to extract basic visual patterns like edges, lines and corners. These extracted features are then combined into the higher layer to form abstract interpretations. On the other hand, shared weights can reduce the complexity of network and avoid overfitting [12].

Rather than investigating features in input space, the proposed algorithm aims at exploring distinctive viewpoints of data through subspace. The Subspace spanned by basis vectors appears as a feature space to capture the most distinguish components of data. To cope with the crucial factors of CNN, basis vectors not only act as transfer functions to access axial components but also receptive fields to extract basic patterns of data. Furthermore, the distribution of a complex real-world data usually cannot be easily described by a single subspace, so multiple-subspace strategy is employed to provide various types of perceptions of data hopefully dealing with intra-class variability.

Inspired by the advantage components of CNN, this study aims at exploring the characteristics of data in subspace manner. Multi-Subspace neural network (MSNN) is proposed to utilize subspace capturing the underlying concept of data distribution. In MSNN, basis vectors trained by adaptive subspace self-organization map (ASSOM) span the subspace and are applied to transfer data from input space to subspace and vice versa. In order to illustrate the proposed network MSNN clearly, four major attributes are listed in the following.

Firstly, basis vectors trained by ASSOM are able to produce distinct filters to ensure small degree of shift, rotation and scaling invariance of input patterns. For instance, translation-invariant filter, rotation-invariant filter and zoom filter are formed while input patterns are randomly displayed. Combining with these qualities embedded in filters makes basis vectors perfect candidates for forming receptive fields. Secondly, basis vectors serve as transfer functions to obtain axial components of data without losing information during delivering data from layer to layer. Thus, basis vectors of ASSOM are capable of preserving information. For the visual task, data has strong two-dimensional relationship. With proper setting of basis vectors, the topology of data could be preserved as well. Thirdly, multiple-subspace strategy is implemented by multiple parallel blocks in MSNN which provide abundant insight of data to deal with intra-class variability even when training samples are few. For the purpose of adjusting these subspaces, ASSOM adopts a competitive learning mechanism to adapt to different underlying characteristics of the training samples. Fourthly, receptive fields defined by basis vectors could extract basic patterns of data. Moreover, the greedy layer-wise unsupervised training scheme is satisfied as well. The chance of vanishing gradient problem is reduced as well [7], [13].

In this study, we have done several experiments to prove the feasibility of the proposed network, MSNN. Experimental results show competitive performance compared to other approaches. The reminder of this study is organized as follows. Section II reviews the structure and learning scheme of ASSOM and CNN. The proposed algorithm MSNN is introduced as well. Datasets and internal organization of network are included in the Section III. Experimental results and discussion are listed in Section IV. Section V summarizes the conclusions of experimental results.

## II. PRELIMINARY KNOWLEDGE

### A. Adaptive Subspace Self-Organizing Map

ASSOM is based on the architecture of self-organization map (SOM) which is first proposed by Kohonen [14]. The structure of ASSOM is shown in Fig. 1. There are three layers in the structure that are similar to neural network, input layer, subspace layer and reconstruction layer. In the input layer, each node represents feature of input data. In hidden layer, a square box, which is referred as a module, represents a single subspace. Nodes in hidden layer indicate basis vectors, which are linearly independent to each other. Since basis vectors span the subspace, number of basis vector should not exceed the dimension of input data.

Let $x \in R^D$ be the input data, $L$ be a subspace spanned by $H$ orthonormal basis vectors $\{b_1^{(L)}, b_2^{(L)} ... b_H^{(L)}\}$, $b_h^{(L)} \in R^D$. So, the projection length $o_h^{(L)}$ on each basis vector is $o_h^{(L)} = b_h^{(L)^T} x$, $for\ h = 1, 2...H$. Final layer, reconstruction layer, is used to estimate the input data. Let $\hat{x}_L$ be the estimation of $x$ and the output of reconstruction layer.

$$\hat{x}_L = \sum_{h=1}^{H} o_h^{(L)} b_h^{(L)} = \sum_{h=1}^{H} b_h^{(L)^T} x b_h^{(L)} \tag{1}$$

-------------------------------------------------- INSERT Fig. 1. --------------------------------------------------

### B. Convolutional Neural Network

CNN has many variant forms and LeNet-5 is one of them designed for handwritten digit recognition. For typical CNN, the network is composed of three types of layer, convolutional layer, pooling layer and fully-connected layer. The structure of LeNet-5 is shown in Fig. 2. LeNet-5 has seven layers excluding input layer. First Layer is a convolutional layer with six feature maps. Second layer is a pooling layer which sub-samples feature maps in first convolutional layer into lower-resolution feature maps. Third layer and fifth layer are convolutional layers. Forth layer is a pooling layer, which has similar effect as second pooling layer. Sixth layer is a fully-connected layer. Final layer is the output layer. After constructing all layers, back propagation is applied to tune the parameters.

-------------------------------------------------- INSERT Fig. 2. --------------------------------------------------

## III. MULTI-SUBSPACE NEURAL NETWORK

This study proposes a network named multi-subspace neural network (MSNN), the structure of which is shown in the Fig. 3. MSNN is constructed by multiple parallel blocks with similar internal organization. Each block is built by inner-product layer, pooling layer, merging layer and fully-connected layer. We will discuss about the formula and meaning of each layer first. Next section will describe the entire learning mechanism.

-------------------------------------------------- INSERT Fig. 3. --------------------------------------------------

### A.  Mechanism for each layer
**(1) Inner-Product Layer**

Due to similar internal organization in each block, we'll discuss one block here. In inner-product layer, there are several feature maps. Feature map indicates the learned representation of input data. Each unit in feature map is the result of neighbouring features (where neighbouring area is demeaned) of the input map inner producing with the kernel. Neighbouring area is defined by the receptive field. With numerous kernels, feature maps consist of different kinds of representation. For the purpose of preserving the information of input data, it is important to define the weights in kernel. In this study, kernels are initialized by basis vectors trained by ASSOM in kernel initialization stage shown in Fig. 4(b).

Each unit in feature map is calculated by inner production operation between kernel and input data, where input data is demeaned.

$$x_j^r = \sum_{u,v} \left[ M_{j_{u,v}}^r - DC_{M_{j_{u,v}}^r} \right] \odot k_j^r, \text{ for } j=1,2...n \tag{2}$$

where $n$ is feature map number in current layer. $x_j^r$ is the jth feature map in rth layer, $M_{j_{u,v}}^r$ is the patch of jth input map. $DC_{M_{j_{u,v}}^r}$ is the DC value of $M_{j_{u,v}}^r$. $k_j^r$ is the kernel between jth input map and jth feature map. Symbol $\odot$ is the inner producting operation.

Kernel acts like a sliding window, which changes its position in input data. Therefore, the associated position of unit in feature map is decided. In the view of mathematic, feature map collects the components of specific axis since kernels are initialized by the basis vectors. Note that parallel blocks construct MSNN, so we'll need multiple subspaces in one time. Basis vectors in one subspace determine kernels in each block.

**(2) Pooling Layer**

Once the feature maps are detected, topology of data has already preserved in feature maps. Therefore, location information becomes less important. Pooling operation is applied to reduce the resolution of feature maps and achieve spatial-invariant. Each feature map in the pooling layer is corresponding to specific feature map in previous layer. Thus, the number of feature maps in current layer is the same as previous layer.

In this study, we employ average-pooling operation to reduce the resolution of feature maps in previous layer.

$$x_j^r = downsample(x_j^{r-1}), \ for \ j = 1, 2...n \tag{3}$$

where $n$ is the feature map number in current layer. $x_j^{r-1}$ is the feature map of previous layer. $x_j^r$ is the jth feature map of current layer, which size is decided by subsampling scale. *downsample* indicates the average-pooling operation.

**(3) Merging Layer**

In the merging layer, there is only one feature map in each block. With an aim of reconstructing data and succeeding the formulas of ASSOM, kernels in the merging layer are identical to kernels in inner-product layer at kernel initialization stage shown in Fig. 4(b). Equation (1) reveals that input data could be restored by summing all the projection lengths multiplying with basis vectors. This procedure is thus implemented by summing feature maps in previous map convoluting with the kernels in current layer for all the feature maps. Convolutional procedure is done by flipping kernel and inner producting kernel with input map. Flipping kernel acts like vector transposed in equation (1), indicating convolution is consistent with meaning of transpose. Then, we add a bias term and pass through activation function. Activation function restricts the outputs of node to specific region and bias term compensates the DC term in the inner-product layer.

$$x^r = f\left[\sum_{j=1}^{n}\left(\sum_{u,v} M^r_{j_{u,v}} \otimes k^r_j\right) + b^r\right] \quad (4)$$

where $x^r$ is the feature map of merging layer. $M^r_{j_{u,v}}$ is the patch of $j_{th}$ feature map of previous layer. $k^r_j$ is the kernel between $j_{th}$ feature map of previous layer and feature map of current layer. $b^r$ is the bias term and $f$ is an activation function. Symbol $\otimes$ indicates convolutional operation. Note that kernels in merging layer would be altered in the back-propagation stage.

Although we desire to rebuild data in the merging layer, convolutional procedure would shrink the size (dimension) of data during reconstruction. Therefore, back propagation is employed to fine-tune the kernels for compensating the effect of dimension shrinking.

**(4) Fully-Connected Layer**

Fully-connected layer usually comes after several inner-product layers, pooling layers and merging layer. Each node (neuron) in fully-connected layer connects all nodes in previous layer. The output of node is computed by summing all the weights multiplying nodes in previous layer and passing through activation function. Equation (5), (6) shows the relation between the nodes in fully-connected layer and layer before that.

$$x^r = f(u^r) \quad (5)$$

$$u^r = W^r x^{r-1} + b^r \quad (6)$$

**(5) Output Layer**

Number of nodes in output layer depends on task. If the network proceeds the classification mission, output node indicates the class label. If the network proceeds the regression mission, output node represents the feature of target.

*B. Learning Scheme of MSNN*

Once initializing all layers in each block, back propagation is employed to fine-tune the parameters for all layers.

The learning mechanism of MSNN is demonstrated by flow chart in Fig. 4(a). Flow chart involves four stages, kernel initialization, network initialization, feedforward, back propagation and orthonormalization stage. Once a training sample goes through five stages, ending criterion will be checked. If ending criterion is satisfied, training process stops. On the other hand, if it doesn't meet the requirement of ending criterion, training process continues. Ending criterion could be fixed epoch number, threshold of mean square error or other requirements. We'll discuss every stage in detail and derive the update rule for network parameters.

-------------------------------------------------- INSERT **Error! Reference source not found.**. -------------------------------------------

**(1) Kernel Initialization**

In first stage, kernel initialization, images are decomposed into small overlapped patches, the size of which is defined by the receptive field. Those patches are firstly demeaned and fed into the training process of ASSOM. After well learning, basis vectors are capable of mapping input data from input space to subspace without losing valuable information. Kernels in the inner-product layer and merging layer are then initialized by basis vectors. Thus, number of basis vector in subspace is equivalent to number of kernel in the inner-product layer, merging layer but it should not exceed the dimension of receptive field. The flow chart of kernel initialization stage is given in the **Error! Reference source not found.**(b).

**(2) Network Initialization**

Second stage, network initialization, randomly defines the value of bias terms in merging layer and connection weights in fully-connected layer.

In addition, there is no limitation about the number of parallel block. During the training process of ASSOM, multiple subspaces are created and compete with each other to get the closest distribution of training sample. Once the training process completes, each subspace has ability to catch the characteristics and distribution of training samples and keeps slight variation from other subspaces. The proposed network takes the advantage of subspace variation to deal with intra-class variability issues.

**(3) MSNN Feedforward**

After initialization, input data would be fed into the network. Input data will go through inner-product layer, pooling layer, merging layer…fully-connected layer and output layer. If the task is classification-oriented, then nodes in output layer stand for the possibility of each class. If the learning process of MSNN is not completed, the output is biased for now.

**(4) MSNN Back Propagation**

In order to link the features with true label, back propagation is employed to fine-tune the parameters in the network including bias terms and kernels in merging layer, kernels in inner-product layer, and weights in fully-connected layer. Our network focuses on two-dimensional data, so there is little difference from ordinary neural network. The error goes from higher layer through lower layer. The derivation will follow the error flow to every layer. Note that the definition of higher layer means layer closest to the output layer while lower layer closet to input layer.

For typical neural network, the cost function for nth training sample is denoted as $E^n = \frac{1}{2}\sum_{k=1}^{c}(t_k^n - y_k^n)^2$, where $t_k^n$ indicates the kth dimension of true output for nth training sample, $y_k^n$ indicates the kth dimension of predicted output for nth training sample and c indicates number of class.

For multi-layer neural network, the output of fully-connected layer is defined by

$$u^r = W^r x^{r-1} + b^r \tag{7}$$

$$x^r = f(u^r) \tag{8}$$

The output is linear combination of input and weight and pass through activation function, where $r$ indicates the current layer, and $x^{r-1}$ is output of (r − 1) th layer, that is to say the input of rth layer. $W^r$ is the connection weights between (r − 1) th layer and rth layer. $b^r$ is a bias in rth layer. $f$ is an activation function.

In order to minimize the cost function, gradient decent algorithm is applied to approach the minimal region. The formula of gradient decent algorithm is defined as

$$W_{new}^r = W_{old}^r - \eta \frac{\partial E}{\partial W_{old}^r} \tag{9}$$

Where $\eta$ is the learning rate.

To derive the update rule for $W$ and $b$, a sensitivity term $\varepsilon^r$ is introduced as the partial differentiation of $u^r$ in rth layer [19].

$$\frac{\partial E}{\partial u^r} = \varepsilon^r \tag{10}$$

The partial differentiation to bias $b^r$ in rth layer is defined by

$$\frac{\partial E}{\partial b^r} = \frac{\partial E}{\partial u^r}\frac{\partial u^r}{\partial b^r} = \varepsilon^r \frac{\partial(W^r x^{r-1}+b^r)}{\partial b^r} = \varepsilon^r \cdot 1 \tag{11}$$

Which substitutes equation (7) into equation (11).

And partial differentiation to connection weight $W^r$ in $r_{th}$ layer is defined by

$$\frac{\partial E}{\partial W^r} = \frac{\partial E}{\partial u^r}\frac{\partial u^r}{\partial W^r} = \varepsilon^r \frac{\partial(W^r x^{r-1}+b^r)}{\partial W^r} = \varepsilon^r x^{r-1} \tag{12}$$

Substituting equation (7) into equation (12) yields the result of partial differentiation to connection weight. From equation (11) and (12), it is obvious that both partial differentiation terms could be represented by sensitivity multiplying constant vector. Sensitivity term could be further rewritten as

$$\begin{aligned}\varepsilon^r &= \frac{\partial E}{\partial u^r} = \frac{\partial E}{\partial u^{r+1}}\frac{\partial u^{r+1}}{\partial u^r} = \varepsilon^{r+1}\frac{\partial(W^{r+1}x^r+b^{r+1})}{\partial u^r}\\&= c\frac{\partial(W^{r+1}f(u^r)+b^{r+1})}{\partial u^r} = \varepsilon^{r+1}W^{r+1}\circ f'(u^r)\end{aligned} \tag{13}$$

Where $\circ$ indicates elementwise multiplication.

Sensitivity term is expressed in a recurrent form which value is determined by sensitivity term from next layer. There is a little variation for sensitivity term in final layer. Here, we assume the output layer, which is denoted as layer *L*, is followed by fully-connected layer. Hence, sensitivity term of final layer is denoted as

$$\begin{aligned}\varepsilon^L &= \frac{\partial E}{\partial b^L} = (y^L - t)\circ \frac{\partial y}{\partial b^L} = (y^L-t)\circ \frac{\partial x^L}{\partial b^L}\\&= (y^L-t)\circ \frac{\partial f(u^L)}{\partial b^L} = (y^L-t)f'(u^L)\end{aligned} \tag{14}$$

So far, we would be able to fine-tune the parameters for every layer. Since the network put emphasize on two-dimensional data, the sensitivity term will be reshaped to two-dimensional shape. Following section will discuss about the amended sensitivity term and update rule for pooling layer, merging layer, fully-connected layer and output layer. Note that the parameters in the inner-product layer remain the same for keeping orthonormal characteristic.

**Output Layer**

The sensitivity term in the output layer is shown in the equation (14). No bias term and connection weights exist in the output layer. Sensitivity term is then transferred into previous layer to update parameters.

**Fully-Connected Layer**

Sensitivity term in the fully-connected layer follows equation (13). The partial differentiate to bias is shown in the equation (11). The partial differentiate to connection weights are shown in equation (12). Once the sensitivity term is calculated, it is back propagated to previous layer for updating all variables.

**Pooling Layer**

Sensitivity term follows equation (13) but should be modified according to the layer following after pooling layer. We will discuss the three cases, fully-connected layer coming after pooling layer, merging layer after it and inner-product layer after it.

If fully-connected layer follows after pooling layer, then $W_j^{r+1}$ in equation (12) indicates the connection weights between two layers.

If merging layer follows after pooling layer. Thus, $W_j^{r+1}$ is replaced by the kernels in merging layer. Sensitivity term is then reshaped into two-dimensional representation, namely sensitivity map. The sensitivity is rewritten as

$$\varepsilon_j^r = \varepsilon^{r+1} k_j^{r+1} \circ f'(u_j^r) = \alpha \circ f'(u_j^r), \text{ for } j=1,2...n, \alpha = \varepsilon^{r+1} k_j^{r+1} \quad (15)$$

where $n$ is the number of kernel in merging layer. $\circ$ indicates elementwise multiplication. By sliding kernel to every pixel of sensitivity map, size of α is identical to the feature map $f'(u^r)$ in current layer.

If inner-product layer follows after pooling layer, then, $W_j^{r+1}$ is replaced by the kernels in inner-product layer. The sensitivity is rewritten as

$$\varepsilon^r = \sum_{j=1}^{n} \varepsilon_j^{r+1} k_j^{r+1} \circ f'(u^r) \quad (16)$$

where $n$ is the number of kernel in inner-product layer. $\circ$ indicates elementwise multiplication. Note that if inner-product layer follows after pooling layer, there is only one feature map in pooling layer.

**Merging Layer**

The sensitivity term follows the equation (13) but should be modified according to the layer coming after merging layer. Three situations are discussed, inner-product layer comes after merging layer, pooling layer after it and fully-connected layer after it.

If fully-connected layer follows after merging layer, $W_j^{r+1}$ in equation (13) indicates the connection weights between two layers. And sensitivity term should be reshaped into two-dimensional representation. If pooling layer follows after merging layer, sensitivity term is rewritten as

$$\varepsilon^r = \varepsilon^{r+1} W_j^{r+1} \circ f'(u^r) = upsample(\varepsilon^{r+1})\beta_j^{r+1} \circ f'(u^r) \qquad (17)$$

where *upsample* indicates that upsampling the sensitivity map in (r + 1)th layer to the same size as previous layer rth. $W_j^{r+1}$ is the subsampling scale.

If inner-product layer comes after merging layer, then sensitivity term is rewritten as

$$\varepsilon^r = \varepsilon^{r+1} W_j^{r+1} \circ f'(u^r) = \sum_{j=1}^{n} \varepsilon_j^{r+1} k_j^{r+1} \circ f'(u^r) \qquad (18)$$

where *n* is the number of kernel in inner-product layer. ∘ indicates elementwise multiplication.

So far, there is everything we require to derive the update rule of parameters. First, we demonstrate the partial differentiation to bias using equation (11).

$$\frac{\partial E}{\partial b^r} = \varepsilon^r = \sum_{u,v}(\varepsilon^r)_{u,v} \qquad (19)$$

Which sums all elements in the sensitivity map as the partial differentiation to bias. Second, the partial differentiation to kernel is shown in the equation (12).

$$\frac{\partial E}{\partial k_j^r} = \varepsilon^r x_j^{r-1} = \sum_{u,v}(\varepsilon^r) \cdot (p_j^{r-1})_{u,v}, \; for \; j = 1, 2...n \qquad (20)$$

Where $(p_j^{r-1})_{u,v}$ is the patch of $x_j^{r-1}$. The partial differentiation of kernels sums all the results of sensitivity maps multiplying with patches of $x_j^{r-1}$. This procedure is implemented by convoluting sensitivity maps with $x_j^{r-1}$.

# IV. EXPERIMENTS – IMAGE DATASET

The proposed algorithm is examined on two benchmark datasets of deep learning, MNIST dataset and COIL-20 dataset. Due to the distinction of dataset, the internal organization of network will be slightly different. Next section will describe two datasets and their internal organization of network that we apply to them. To be simply, inner-product layer is labelled $\alpha IP\beta$, pooling layer as $P\gamma$, merging layer as $\alpha M\beta$ and fully-connected layer as $F\varphi$ in the following. Where $\alpha$ is the number of feature map in the layers, $\beta$ is referred as size of kernel, $\gamma$ is the pooling scale and $\varphi$ is the number of node. Note that experiments are performed at 64 bits Windows 7 operating system with Intel core i5 and 10GB RAM. Program is implemented using MATLAB R2014b version.

## A. MNIST Dataset

MNIST dataset [6] collects lots of handwritten digits between numerical 0 to 9. Those digits have been centered in the fixed-field and normalized into 28×28 greyscale pixel. MNIST dataset includes a training set with 60,000 samples and test set with 10,000 samples. Each category is almost equally distributed in both train and test set.

The network is designed at the form, 24 parallel blocks with 28×28-10IP5-P2-1M5-P2-F280 internal organization. Kernels in the network are first initialized by basis vectors trained by ASSOM. There are 24 competitive modules during the training process of ASSOM, which are trained for 15 epochs, exponential learning rate at initial value 0.5 and decay factor 0.05. Network is then trained for 200 epochs with 50 batch samples, exponential learning rate sets at initial value 1 and decay factor 0.005.

Note that number of parallel block is determined by validation test and dominates how many the competitive modules will exist during the training process of ASSOM. The competitive modules must be more than parallel block number for the purpose of initialization concern.

Input layer represents an input image with size 28×28.

First layer, 10IP5, is an inner-product layer with 10 feature maps of size 24×24. Each unit in the feature map connects a 5×5 neighbouring area of input map. In other words, receptive field is 5×5 and there are 10 kernels of size 5×5. Kernels are defined by ASSOM. How many numbers of kernel existing in the network is related to the basis vectors in the module of ASSOM. For 25-dimensional data, 10 basis vectors are enough for the subspace to approximate the original data. If dimension of input data is higher than 25, it will require more than 10 basis vectors to approximate.

Second layer, P2, is a pooling layer with 10 feature maps of size 12×12. Each unit in the feature map connects a 2×2 neighbouring area of feature map from inner-product layer. The 2×2 neighbouring area is not overlapping. Therefore, the size of feature map in current layer is half of feature map in previous layer.

Third layer, 1M5, is a merging layer with one feature map of size 8×8. Each unit in the feature map connects every 5×5 neighbouring area of feature map in pooling layer. Size of receptive field is decided according to the inner-product layer. Kernels in current layer are initialized as kernels in the inner-product layer to meet the requirements of ASSOM formula. In the back-propagation process, kernels will be further fine-tuned.

Forth layer, P2, is pooling layer with one feature map of size 4×4. Each unit in the feature map connects a 2×2 neighbouring area of feature map of merging layer. The procedure is the same as second layer.

Fifth layer, F280, is fully-connected layer with 280 nodes. Nodes in fully-connected layer connect every node in previous layer.

Output layer contains 10 nodes according to the number of category in MNIST dataset. For multiclass classification problem, the target label is usually organized as one-of-c rule. For example, $h_{th}$ element will be one and rest of elements will be zero in label pattern if the sample belongs to class h.

## C.  COIL-20 Dataset

Columbia Object Image Library (COIL-20) [20] is a database which collects 20 objects and 72 images per object. Duck, truck, cup…are included in the category. Objects are taken at every five-degrees rotation. There are 72 images per object and 1440 images in total. In this experiment, we use pre-processed images that are centered in the field and background has been removed. Image size is 32×32 greyscale pixel with intensity normalized into the range [−1, 1].

The network is designed at the form, 16 parallel blocks with 32×32-10IP5-P2-1M5-P2-F250 internal organization. Kernels in the network are first initialized by basis vectors trained by ASSOM. There are 16 competitive modules during the training process of ASSOM, which are trained for 10 epochs, exponential learning rate at initial value 1 and decay factor 0.05. Network is then trained for 400 epochs with 5 batch samples, exponential learning rate set at initial value 0.5 and decay factor 0.005.

# V. RESULTS AND DISCUSSIONS

For the purpose of verifying the feasibility of the proposed algorithm, we examine the algorithm on two benchmark datasets of deep learning, MNIST dataset and COIL-20 dataset. COIL-20 dataset doesn't split the dataset into train set and test set. 50 samples per category (1,000 labelled samples in total) are taked as train set and rest of samples (440 samples) as test set in the following experiment.

we first investigate whether the network will improve or decline with the cooperation of basis vectors. At the following, we compare classification performance with six well-known algorithms, convolutional neural network (CNN), deep belief network (DBN), multi-layer neural network (ANN), PCA network (PCANet), LDA network (LDANet) and random network (RandNet). Wrong predicted samples will be shown to probe the potential reasons that cause it. Furthermore, we are interested in what features and kernels are learned when the network is well-trained. Figural changes are displayed in terms of layers. Visualizing the data helps us to identify whether the information will lose or distort during transmitting data from layer to layer.

At the end, we'll discuss about the influence of parameters. For example, kernel size, kernel number, block number…are included as well.

## A. *Does Basis Vector Help the Optimization of Deep Network?*

Recently, there are several studies [7], [21] performing empirical experiments to prove that deep network which experiences a difficult non-convex optimization problem with greedy layer-wise pre-training does help the optimization in generalizing performance on test set, holing a start point for supervised fine-tuning.

Based on those incredible findings of previous studies, our proposed network employed subspace concept in kernels to migrate the deep network into a local minimum region. In order to evaluate the degree of improvements, we did the experiment on the MNIST dataset and COIL20 dataset with the help of subspace and without subspace.

TABLE I reveals the classification error rate with and without subspace embedded in kernels on training samples. The best result is highlighted in bold-face type. From TABLE I, network with kernels defined by basis vectors performed better than kernels defined by random variables which indicated introducing subspace in the deep network did improve the performance of deep network.

However, in COIL-20 dataset, MSNN with kernels initialized by basis vectors performs worse than kernels initialized by random variables. Since the network has lots of parameters to be trained, overfitting may occur if training samples are scarce [11], [12].

------------------------------------------- INSERT TABLE I -------------------------------------------

### B. Comparison of Classification Performance

TABLE II summarizes the classification test error of LeNet-5, 2-NN, DBN, PCANet-1, LDANet-1, RandNet-1 and MSNN on MNIST dataset. The best result is highlighted in bold-face type. Results of LeNet-5 are taken from LeCun et al. 1998 [6], 2-NN are from Weston et al, 2008 [22], DBN are from Hinton et al, 2006[7], PCANet-1, LDANet-1 and RandNet-1 are from Chan et al, 2015 [1]. From TABLE II, PCANet-1 achieves 0.94% test error rate, outperforming than other algorithms and our proposed network MSNN shows identical performance with LeNet-5 at 0.95% test error rate.

TABLE III gives the classification error on deep learning algorithms using COIL-20 dataset. The cross-validation sets CNN as six-layer structure (C6-S2-C16-S2-F400-output layer), 3-NN as three-layer structure with 200 nodes, 100 nodes in first and second hidden layer, DBN as two-layer structure with 500 nodes in hidden layer and MSNN as six-layer structure (IP10-P2-M1-P2-F250-output layer). In order to keep fair comparison, the patch size and overlap ratio in PCANet-1, LDANet-1, and RandNet-1 are 5 by 5, 0.8 respectively which is identical to the setting of MSNN. Best recognition error rate is obtained by CNN and PCANet-1 at 0%. MSNN achieves 0.09% error rate which comes from merely one misjudged sample. We will take a close look at one misjudged sample in next section.

--------------------------------------------- INSERT TABLE II ---------------------------------------------
--------------------------------------------- INSERT TABLE III---------------------------------------------

### C. Error Discussion

To find the potential reasons that cause bad judgments, we pull out one record of classification result and analyze its confusion matrix. Fig. 5(a)(b) are the confusion matrixes for test set and train set.

Rows of confusion matrix correspond to true label and columns of which correspond to predicted label. Diagonal indicates right judgments and rest of matrix indicates wrong predicted samples. From Fig. 5(a)(b), digits (2, 7), digits (3, 5), digits (4, 9) are pairs suffering distinguished issue. Digits (2, 7), digits (3, 5), digits (7, 9) make most of mistakes. Similar result can be also observed. Those training digits with bad judgments usually had bad performance on testing experiment as well. Fig. 5(c) shows the wrong predicted samples of test set.

The notation α→β on the top of digit indicates the true and predicted label, where α is the true label and β is predicted label. Let us pay attention at the samples where digit seven is mistaken for digit two. Those samples all have horizontal line passing through the body of digit seven and may further be considered as the bottom horizontal line of digit two. Conversely, those samples which digit two is mistaken for digit seven almost depicts the semi-circular curve of digit two as horizontal line. It may confuse with the top horizontal line of digit seven. Taking a close look at wrong predicted samples, these samples often have ambiguous strokes that may confuse the classifier.

As for COIL-20 dataset, the error for MSNN is caused by one wrong predicted sample, image from $19_{th}$ category being mistaken for first category. Fig. 5(d) demonstrates the one wrong predicted sample.

--------------------------------------------------- INSERT Fig. 5. ---------------------------------------------------

### D. Graph Visualization

In this section, we are going to visualize the feature maps distributed in all layer and kernel maps in the inner-product layer and merging layer. Visualizing the feature maps helps us to identify whether the information will lose or distort during transmitting data from layer to layer. Moreover, it is a better way to understand the influence that layer has bought us.

Digit two and seven are chosen to display the formed feature maps. Input layer displays the input image. First layer is inner-product layer. Kernels in this layer are initialized as basis vectors of subspace, thus each feature map can be illustrated as the components of specific dimension. The first few basis vectors usually dominate the rough contour of image, and resting of them relates to elaborate details of graph. Therefore, the last few feature maps are crumbling. Removing the last few feature maps had minor influence on reconstructing data but the first few ones does a great influence. Second layer is pooling layer. We apply averaging pooling to reduce the resolution of feature maps from previous layer. So, feature maps are coarser than previous layer. Third layer is merging layer. Judging from the feature map, we can still distinguish which digit it is despite of limited resolution. Conservative speaking, we can claim that MSNN has ability to reduce the dimension of input data while preserving most of valuable information including topology of data.

Fig. 6(a)(b) demonstrates the kernel maps of inner-product layer and merging layer at the first parallel block of network. Kernel maps will reflect the characteristic of input data [14]. For instance, kernels are formed with distinctive graph according to the input pattern, which possesses small degree of translation, rotation or scaling property. Fig.

6(c) shows the feature maps of digit two at the first block of network. It is clear that data topology is preserved even layer-to-layer transmission.

-------------------------------------------------- INSERT Fig. 6--------------------------------------------------

### E.  *Parameter Analysis*

In this section, the effects for network parameter are discussed. Since we focus on network parameters, we apply 1,000 training samples with equally distributed class on the following experiments. Three parameters, kernel size, kernel number, block number are included in the content. Classification performance and figural changes are listed as well.

**(1) Kernel Size**

Fig. 7(d) lists the classification performance of MNIST dataset with respect to kernel size 3×3, 5×5 and 7×7 pixel. In this experiment, parameters except kernel size in the network are identical to maintain the fair environment. Kernel size 5×5 is most suitable for MNIST dataset. Kernel size 3×3 and 7×7 also reveals rather bad performance.

Fig. 7(a) shows the feature maps in terms of different kernel sizes on digit five and six. If kernel size is 3×3, the feature maps turn out to be more delicate than that of which kernel size is seven in merging layer. The reason behind it can be explained mathematically. We can imagine graph as complicated function. Any curve or high order function could be approximated with the lines. The concept is that curve can be decomposed into small pieces. With small enough piece, pieces look like lines. Pieces in this subject refer to kernel map. With smaller size of kernel map, the trivia parts of input have bigger opportunity to be preserved.

As for COIL-20 dataset, Fig. 7(g) gives classification test error with respect to kernel size 3×3, 5×5 and 7×7 pixel. Kernel size 5×5 has superior performance at 0.09% error rate than other size of kernel size.

**(2) Kernel Number**

Fig. 7(e) lists the classification performance of MNIST dataset in terms of kernel numbers 10, 15 and 20, using in inner-product and merging layer. From Fig. 7(e), classification performance improves when the kernel number is increased. Kernel numbers refer to the number of basis vector in the module of ASSOM. During the training process of ASSOM, images are decomposed into small overlapped patches with identical size as kernel. Patches are collected for training the basis vectors of competing modules. The number of basis vector relates the degree of similarity between reconstructing data and original data. For 25-dimensional data, 10 basis vectors are enough for recovering. Data reconstructed with 15 and 20 basis vectors is more like to the original one but difference is little. Since the last few basis vectors dominate the delicate parts of input data, there is no much variation for restring data in visual presentation. Fig. 7(b) shows the feature maps formed with different kernel number on digit five and six.

As for COIL-20 dataset, Fig. 7(h) gives three settings of kernel number. Classification error reached its low point at 0.09%.

**(3) Block Number**

Fig. 7(f) lists the classification performance of MNIST dataset with respect to 10, 16 and 20 parallel blocks. Each block has the same structure but with different initial value, determined by the kernel initialization stage. More blocks in the network indicate more information is provided to classifier, so network performance assumes to be better while increasing block number. This assumption is proved when the network is trained with entire training samples of MNIST. But for 1000 training samples, classification performance achieves 5.84% error rate at 16 numbers of block but decays when increasing to 20 numbers of block. The reason may be 1000 training samples are not enough for the deep network to well train every connection. Once reaching the critical point of block number, the performance is going poor. Fig. 7(c) shows the feature maps of digit five and six at the first, fifth and tenth block. Feature maps at different block share similar contour but exist slightly variations.

As for COIL-20 dataset, Fig. 7(i) gives the classification error with respect to number of parallel block. Network with 10 block number does better than other number and reaches 0.64% error rate.

Note that ASSOM dominates the kernel initialization stage and then utilizes the learned basis vectors of competing modules to define the kernel. The reason why we employ parallel block is taking advantage of competing modules. Competing modules grow to model input data. At the end of training process of ASSOM, competing modules possess analogous characteristics of data but preserve slight difference. We expect that slight difference could be against the noise from input data, namely intra-class variability issues. Exact number of block is determined by cross-validation test.

--------------------------------------------------- INSERT Fig. 7--------------------------------------------------

*F. Noise Sensitivity*

From TABLE III, the classification performances of CNN, PCANet, LDANet, RandNet and MSNN are almost identical. Therefore, we have done noise sensitivity experiments on weights to prove which network is more robust than others. To be fair, since the MSNN focuses on two-dimensional data, we only choose those networks which target on two-dimensional data, such as CNN, PCANet, LDANet and RandNet. In the following experiments, Gaussian noises with zero mean and 0.5, 0.75 1 standard deviation are generated and added on the kernels of networks after training process is completed. Noted that experiments are to model the networks are affected by the noise. Not all the kernels will be influenced by noise. Hence, Gaussian noises will be imposed on all kernels but only partial weights of kernel will be disturbed. The proportion of affected weights in kernel is called noise level. We have 10%, 20% and 30% noise level in the experiments. Note that the setting of all networks is identical. N ($m$, $s$) indicates Gaussian noise with $m$ mean and $s$ standard derivation. The best result is highlighted in bold-face type.

TABLE IV reveals the testing error rate of 10%, 20% and 30% Gaussian noises with zero mean and 0.5, 0.75, 1 standard derivation imposed on the kernel weights. As illustrated by those tables, the performance of all networks is getting awful while standard derivation of noise increases. In the conditions of N (0,0.75) and N (0,1), the MSNN beats other approaches except 30% N (0,1). From the results, the MSNN has lower noise sensitivity than others, which implies the MSNN is robust even when network is disturbed by noises.

-------------------------------------------------- INSERT TABLE IV--------------------------------------------

## VI. CONCLUSIONS AND FUTURE WORK

In this study, we have presented a network MSNN that utilize subspace concept capturing the underlying concept of data distribution for image classification. Basis vectors trained by ASSOM are spanning the subspace and applied to transfer data from input space to subspace without losing original information. Moreover, MSNN integrates the key factor of CNN so basis vectors not only serve as transfer functions to access axial components but also receptive fields to extract basic patterns of data. Considering real-world scenario, data cannot be easily described by a single subspace, so multiple-subspace strategy is implemented as multiple parallel blocks structure to provide various viewpoints of data dealing with intra-class variability issues. By doing this, there are four major benefits in the following.

Firstly, basis vectors trained by ASSOM are capable of producing distinct filters to ensure small degree of variance. Secondly, basis vectors act as transfer function to deliver data from layer to layer without losing information and distorting the data topology. Thirdly, multiple parallel blocks provide various, high-level and abstract insights of data which make the network more robust. Fourthly, receptive fields initialized by basis vectors could extract basic patterns of data and form complicated graph at higher layer. Furthermore, the greedy layer-wise unsupervised training scheme is satisfied by initializing the value near the local minimum region. Such strategy could alleviate the chance of vanishing gradient problem in deep network [13].

Our experiments showed the performance of proposed network was better than the network without subspace concept if training samples are sufficient. We also conducted the experiments on two benchmark datasets of deep learning, MNIST dataset and COIL-20 dataset. The experimental results revealed that the proposed network achieves competitive performance compared to current state-of-the-art approaches. Furthermore, for the purpose of evaluating the robustness of network, noise sensitivity experiments are conducted on COIL-20 dataset. Experimental results imply that the structure of MSNN makes it more resistant to noises.

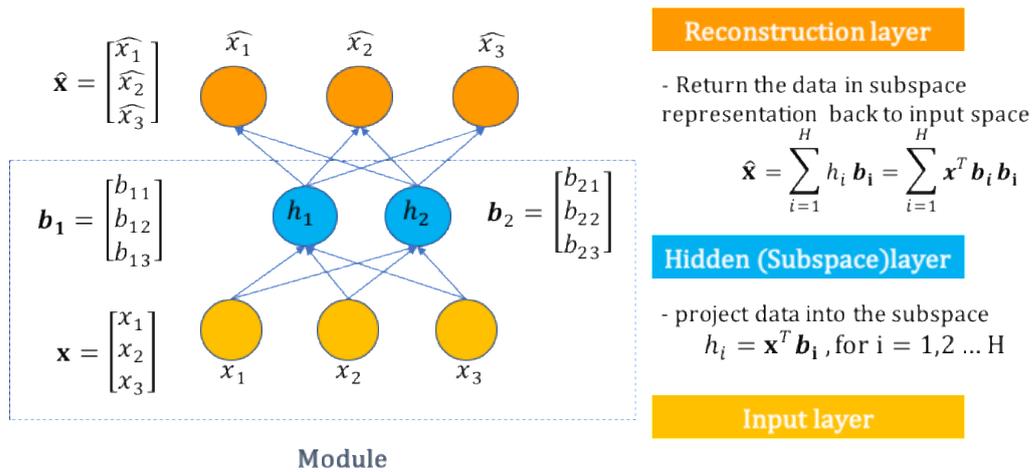

Fig. 1. Structure of ASSOM

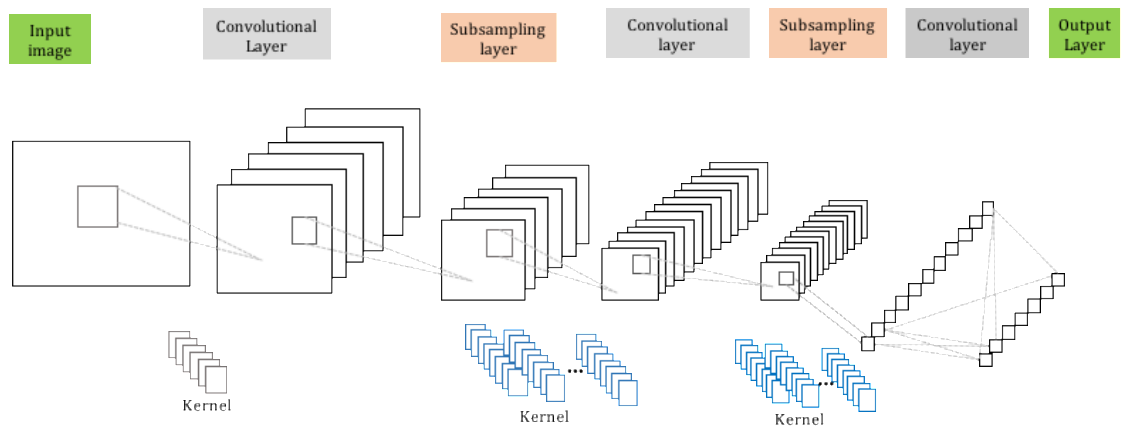

Fig. 2. Structure of LeNet-5

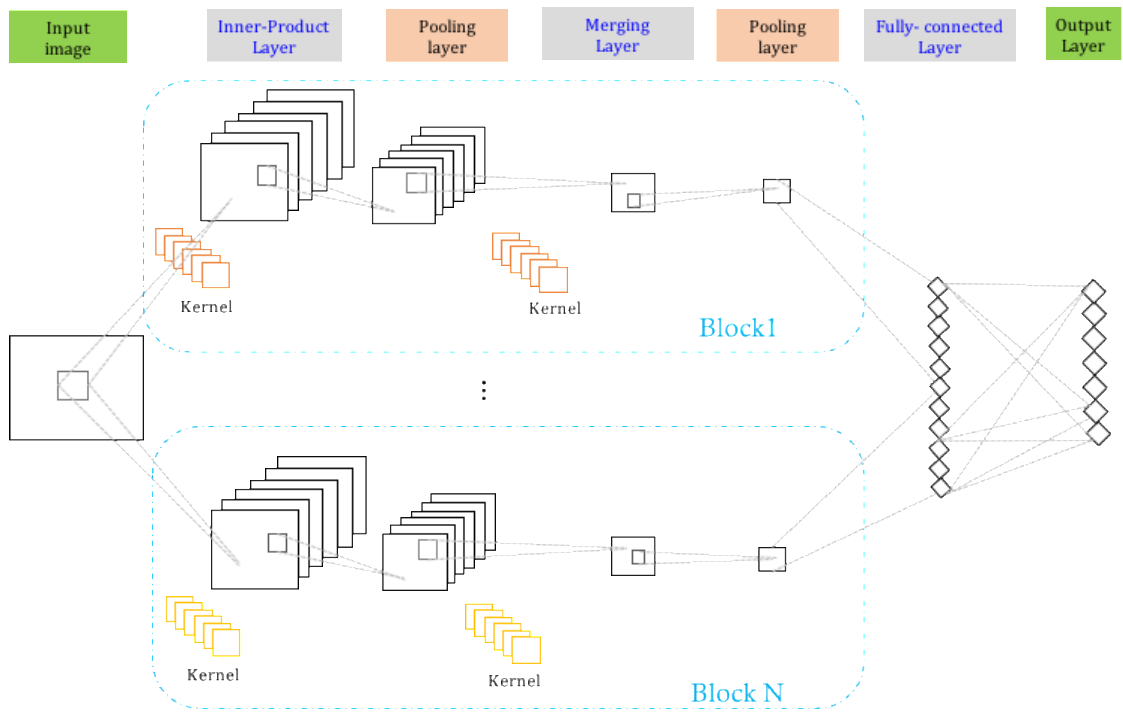

Fig. 3. Structure of MSNN

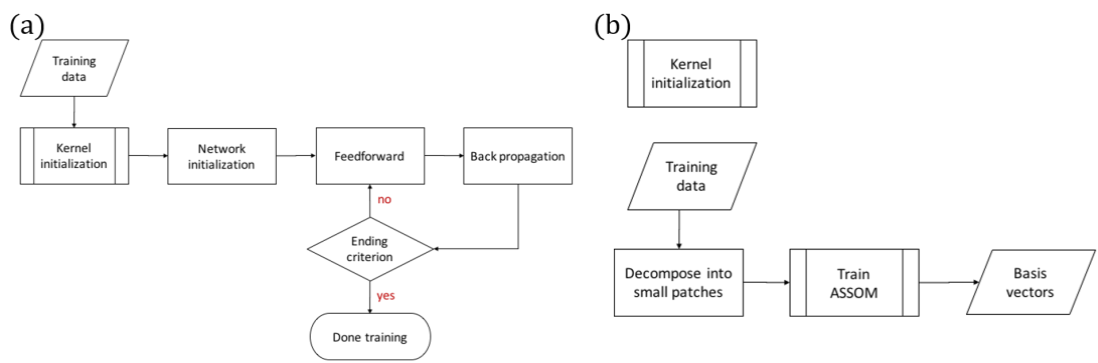

Fig. 4. (a) shows the block diagram of MSNN learning scheme and Fig. (b) shows the diagram of kernel initialization stage.

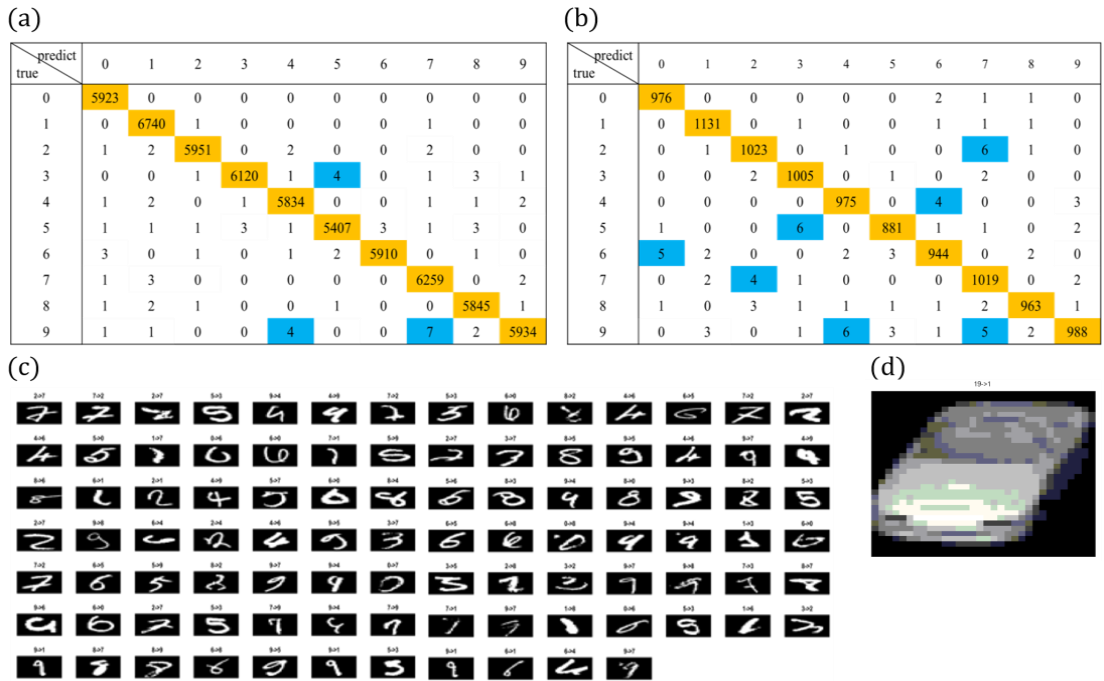

Fig. 5. (a)(b) are the confusion matrixes of MNIST train set and test set. (c) collects the misclassified samples of MNIST test set. (d) is the wrong predicted sample of COIL-20 dataset.

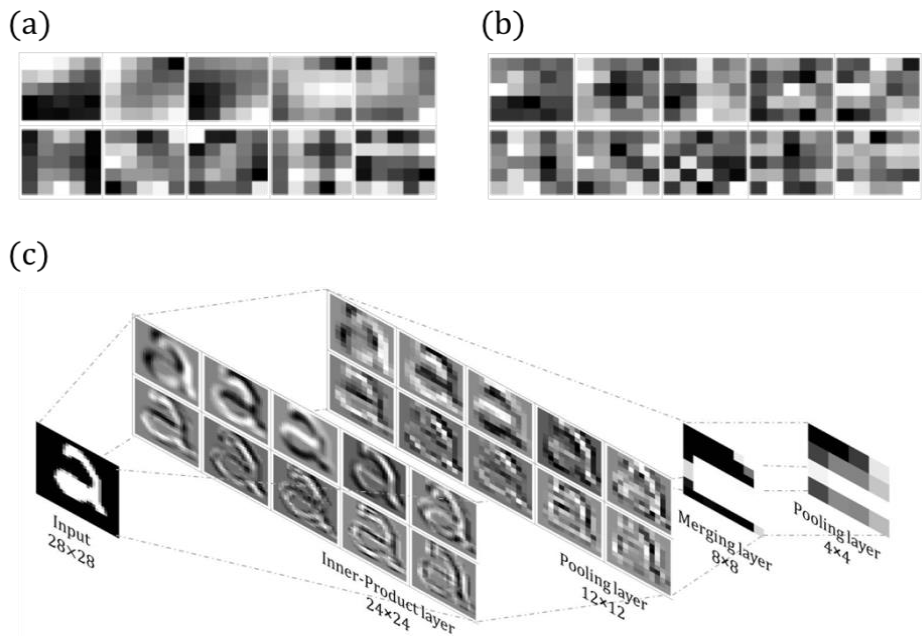

Fig. 6. (a) shows the learned kernels of inner-product layer at the first block of network. (b) shows the learned kernels of merging layer at the first block of network. (c) shows the feature maps of digit two at the first block of network.

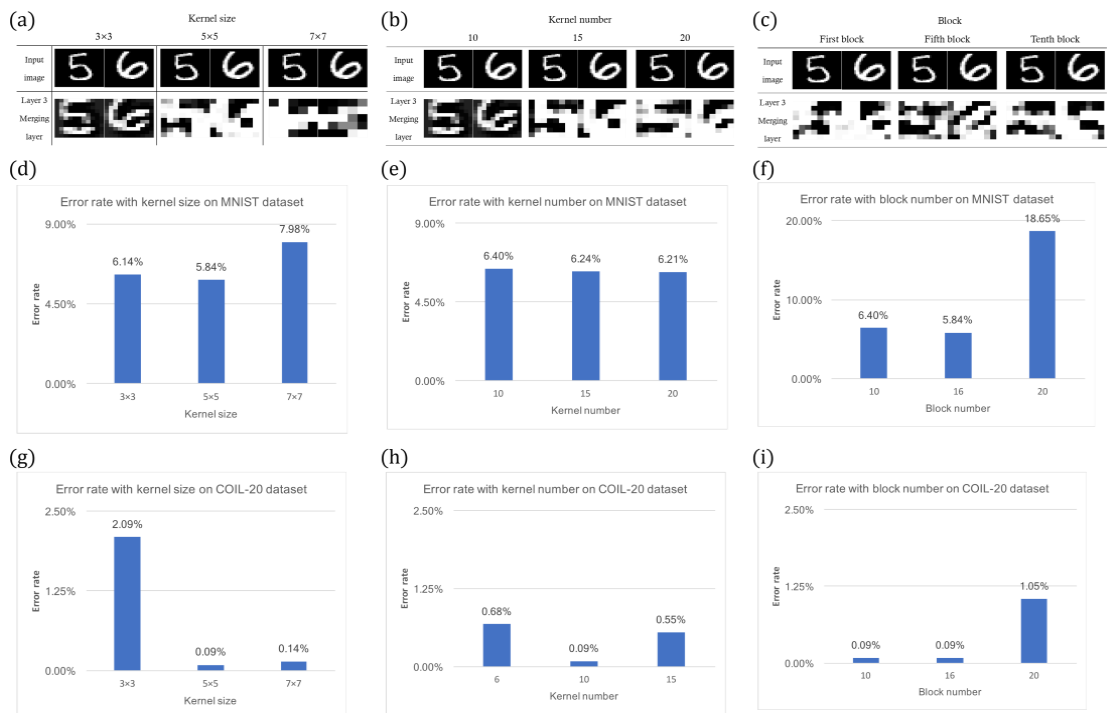

Fig. 7. (a) shows the feature maps of digit five/six in terms of 3×3, 5×5 and 7×7 kernel map. (b) shows the feature maps of digit five/six in terms of 10, 15 and 20 kernel number. (c) shows the feature maps of digit five/six in terms of first, fifth and tenth block of network. (d) is the error rate of MNIST dataset in terms of kernel size. (e) is the error rate of MNIST dataset in terms of kernel number. (f) is the error rate of MNIST dataset in terms of block number. (g) is the error rate of COIL-20 dataset in terms of kernel size. (h) is the error rate of COIL-20 dataset in terms of kernel number. (i) is the error rate of COIL-20 dataset in terms of block number.

TABLE I

TESTING ERROR RATE WITH AND WITHOUT SUBSPACE EMBEDDED IN KERNELS

| Algorithm | MNIST | COIL-20 |
| --- | --- | --- |
| MSNN with kernels initialized by basis vectors | **1.10%** | 0.64% |
| MSNN with kernels initialized by random variables | 1.22% | **0.36%** |

TABLE II

TESTING ERROR RATE OF DEEP LEARNING ALGORITHMS ON MNIST DATASET

| Algorithm | Testing error rate |
|---|---|
| LeNet-5    (LeCun et al. 1998) | 0.95% |
| 2-NN    (Weston et al., 2008) | 1.60% |
| DBN    (Hinton et al., 2006) | 1.17% |
| PCANet-1 (Chan et al., 2015) | **0.94%** |
| LDANet-1 (Chan et al., 2015) | 0.98% |
| RandNet-1 (Chan et al., 2015) | 1.32% |
| MSNN | **0.95%** |

TABLE   III

ERROR RATE OF DEEP LEARNING ALGORITHMS ON COIL-20 DATASET

| Algorithm | Testing error rate |
|---|---|
| CNN | **0.00%** |
| 2-NN | 0.95% |
| DBN | 0.72% |
| PCANet-1 | **0.00%** |
| LDANet-1 | 0.04% |
| RandNet-1 | 0.05% |
| MSNN | 0.09% |

TABLE   IV

ERROR RATE OF 10%, 20% and 30% GAUSSIAN NOISE IMPOSED ON KERNEL WEIGHTS ON COIL-20 DATASET

|  | Algorithm | N (0,0.5) | N (0,0.75) | N (0,1) |
|---|---|---|---|---|
| 10% Gaussian Noise Level | CNN | 0.59% | 1.41% | 8.09% |
|  | PCANet-1 | 8.41% | 16.68% | 20.73% |
|  | LDANet-1 | 30.23% | 50.59% | 16.14% |
|  | RandNet-1 | **0.14%** | 8.14% | 13.68% |
|  | MSNN | 0.32% | **1.09%** | **3.00%** |
| 20% Gaussian Noise Level | CNN | 0.55% | 6.23% | 13.68% |
|  | PCANet-1 | 22.95% | 11.41% | 36.18% |
|  | LDANet-1 | 20.09% | 41.77% | 45.41% |
|  | RandNet-1 | **0.41%** | 13.09% | 6.32% |
|  | MSNN | 0.77% | **3.14%** | **6.18%** |
| 30% Gaussian Noise Level | CNN | 22.45% | 22.73% | 30.68% |
|  | PCANet-1 | 24.77% | 16.55% | 26.36% |
|  | LDANet-1 | 39.05% | 39.32% | 67.0% |
|  | RandNet-1 | **0.68%** | **0.23%** | 7.23% |
|  | MSNN | 1.45% | 3.96% | **7.09%** |